\documentclass{article}

\usepackage{ijcai17}

\usepackage{graphicx}
\usepackage{array}
\usepackage{multirow}
\usepackage{multicol} 
\usepackage{mathrsfs}
\usepackage{algorithmic}
\usepackage[inoutnumbered,linesnumbered,ruled,vlined]{algorithm2e}
\usepackage{epsfig}
\usepackage{epstopdf}
\usepackage{subfigure}
\usepackage{enumerate}
\usepackage{times}
\usepackage{helvet}
\usepackage{courier}
\usepackage{url}
\usepackage{amsmath,amssymb,amsthm}
\usepackage{color}
\usepackage[small]{caption}
\usepackage{xfrac}
\usepackage{bbding}
\usepackage{pifont}
\usepackage{rotating}
\usepackage[table]{xcolor}% http://ctan.org/pkg/xcolor
\usepackage{mathtools}
\usepackage{booktabs}
\newcommand{\skipNow}[1]{}

\DeclareMathAlphabet{\mathpzc}{OT1}{pzc}{m}{it}
\DeclareMathAlphabet\mathbfcal{OMS}{cmsy}{b}{n}

\begin{document}

% \title{Interpreting Random Forests in the Semantic Web}
% \title{Interpreting Classification in the Semantic Web}
% \title{Towards Ontology Stream Learning for Consistent Prediction}
%\title{Learning from Ontology Streams with Semantic Concept Drift\\\footnote*{\small{Preprint of paper accepted at IJCAI 2017 - to be presented at Melbourne, Australia in August 2017}}}
\title{Local Score Dependent Model Explanation for Time Dependent Covariates\thanks{Work accepted as full paper for presentation at XAI (Explainable AI) workshop at Twenty-Eighth International Joint Conference on Artificial Intelligence (IJCAI) 2019 in Macao, China -  August 10-16, 2019.}} 

% \author{
% \# 2411

\author{
Xochitl Watts\\
Chicago, USA\\
xwatts@gmail.com
\And
Freddy L\'ecu\'e\\
Inria, France\\
CortAIx @Thales, Montreal, Canada \\
freddy.lecue@inria.fr
}

\maketitle

\begin{abstract}
The use of deep neural networks to make high risk decisions creates a need for global and local explanations so that users and experts have confidence in the modeling algorithms. We introduce a novel technique to find global and local explanations for time series data used in binary classification machine learning systems. We identify the most salient of the original features used by a black box model to distinguish between classes. The explanation can be made on categorical, continuous, and time series data and can be generalized to any binary classification model. The analysis is conducted on time series data to train a long short-term memory deep neural network and uses the time dependent structure of the underlying features in the explanation. The proposed technique attributes weights to features to explain an observations risk of belonging to a class as a multiplicative factor of a base hazard rate. We use a variation of the Cox Proportional Hazards regression, a Generalized Additive Model, to explain the effect of variables upon the probability of an in-class response for a score output from the black box model. The covariates incorporate time dependence structure in the features so the explanation is inclusive of the underlying time series data structure. 
\end{abstract}
\vspace{-0.2cm}

\section{Introduction}
Machine learning models are broadly used as a tool for classification problems, and deep neural networks are common because they achieve high predictive accuracy although they are opaque classifiers. We are focused on explaining a model decision caused by the feature inputs when the features are time series data and the black box model is a recurrent neural network. 

Explanation of time series classification is  differentiated because the attributes are ordered [Bagnall et al.; 2016]. While there has been no formal or technical agreed upon definition of model explanations; explanations for machine learning models provide an account that makes the model classification decision more clear. Explanations describe the decision made by a machine learning model in order to gain user acceptance and trust [Lipton, 2016; Weld and Bansal, 2018], legality purposes that come from ethical standards and the right to be informed about the basis of the decision [Goodman and Flaxman, 2016; Wachter, 2017], debugging the machine learning system to identify flaws and inadequacies or distributional drift [Kulesza et al., 2014; Weld and Bansal, 2018], and lastly for an increase in insight to the domain area for instance uncovering causality [Lipton, 2016]. We are interested in post hoc model explanation to determine how the classification model behaved and why through global explanations for the entire decision space as well as local explanations for observations within a region of the model score output. 

To our knowledge, our work provides the first time that the model score and response has been proven to be a Markov process state space model and the explanation incorporates time dependent data for global explanations and score dependent explanations. We begin the analysis using the Product Limit Estimator to derive a non-parametric statistic used to estimate the cumulative probability of an observation being a true responder over the black box model score. The explanation method then uses a semi-parametric model, Cox Proportional Hazards (CPH) regression, on the cumulative probability curve to explain the model scores using the model attributes. The explanations are extended to incorporate time dependent covariates and score dependent coefficients with a Generalized Additive Model (GAM). The application we describe can be applied generally where the features used in the black box model have a causal relationship to the classification label. We provide experimental evidence for time series data to use CPH as an explanation, but are limited in explaining certain models and datasets due to the curse of dimensionality where a dataset requires more true positive observations than the number of covariates used in the explanation. 

\section{Related Work}

\subsection{Recurrent Neural Networks}
Recurrent neural networks (RNN) are a black box model where it is unclear how their decision output is produced. We use long short-term memory (LSTM) cells [Hochreiter and Schmidhuber, 1997] that allows for sequences of data to be processed and keeps a hidden state over the series in the sequence. The LSTM cell uses gating mechanisms to read from, write to, or reset the cell. It is well-suited for classifying time series data and mitigates the vanishing gradient problem in RNNs [Bengio et. al, 1994]. The network learns a dense black-box hidden representation of the sequential input and is able to classify time series data. Our LSTM network also includes nonlinear deep fully connected layers which are unexplainable hidden layers. 

\subsection{Explanation of Recurrent Neural Networks}
Explanations for time series neural networks is addressed at a global level to visualize the internal state of the model but also at a local level which combines visualizations with experimentation of the feature inputs to understand the prediction for a single observation. Visualizations of the changes of the internal state of the model over sequences of input aim to identify patterns where the model has learned hidden state properties of the dataset [Strobelt et al., 2017] and allows for what-if style exploration of trained models [Strobelt et al., 2018]. Another form of global and local explanations are learned prototypes generated from the latent space of the model which can be visualized on projections onto reduced dimensionality and learned prototypes are capable of learning real-world features [Gee et al., 2019]. Sensitivity analysis [Tabatabaee et al., 2013] varies features over a range of values to determine the variance in predicted values by features. Deep Taylor decomposition [Montavon et al., 2015] redistributes relevance onto features which are fed into the model to produce a heatmap of relevant inputs.

\section{Background}{\label{sec:Background}}
The method is derived using the assumption of an underlying Markov process and methods developed in the field of Survival Analysis. The stochastic counting process is used to derive a Product Limit Estimator to derive a non-parametric statistic used to estimate the cumulative probability of an observation being a true responder over the black box model score.

The state space of an observation in the binary classification model has a cardinality of 3 and is either a responder, a non responder, or unknown response. The responder is an observation in the class modeled, a non responder is out of the class, and an unknown response is censored where the value of the observation is only partially known or it is an unlabeled observation. We view an observation moving from the non responder state to a responder state as a stochastic process. An individual observation moves from one state to another due to observation factors which are used as inputs to the black box classification model – therefore this technique can only be used where there is evidence of cause and effect. Furthermore, the non responders are truncated from the analysis because the state is absorbing where it is impossible to go from a non responder to a responder given any feature set and are not part of a stochastic process. Unlabeled observations provide some information with the model score output and we incorporate these observations as censored data. 

We can now formulate an adaptation of a Markov model as a convenient and intuitive tool for constructing hazard models for a response to occur at a certain score interval. The Markov process and the Martingale process are used for simplifying the dependence structure of a stochastic process. The stochastic process is an observation changing from a non responder to a responder over the indexed value of the model score. The index set used to index the random variable in this adaptation is the score output from the binary classification model rather than using time in survival analysis or traditional Markov processes as the index set. Model score is an ordered sequence and is analogous to time. We define the concepts of a past and a future in terms of lower or higher score.

The Markov definition is a simplification of the transition probabilities that describe the probability for the process to move from one state to another within a specified score interval. A Markov process is memoryless [Paul and Baschnagel, 2013], once we know the current state of the process, any knowledge of the “past” or any circumstance in which an observation receives a lesser score does not give any further information about the state of the process in the “future” or in this case a higher score. It suffices to make use of the current state to describe the probability distribution of the process over the score interval [Paul and Baschnagel, 2013].

\section{Theoretical Analysis}
The random process of responder observations over the model score can be modeled as a Markov process. The Markov model is able to describe the risk process of a responder observation at a score that is output from the black box classification model. We say that $X(s)$ is Markov if
$
P(X(s) = x |X(s_k) = x_k, X(s_{k-1}) = x_{k-1},..., X(s_1) = x_1) = P(X(s) = x|X(s_k) = x_k)
$
for any selection of score points $s_1, ..., s_{k-1}, s_k$ such that $s_1 < ... < s_{k-1} < s_k $ and integers $ x_1, ..., x_{k-1}, x_k$. The assumption holds as long as the value of $X$ at lower scores is uninformative when predicting outcomes of $X$ at higher scores, or lower scores and higher scores are independent given the score $s$. The Markov property is time-homogenous when the transition probabilities only depend on the score $s$ and not on the starting score. A set of input values may change over time for an observation where a model score is output given the current set of feature inputs is independent of time points. 

The following definitions are derived from parallel survival analysis equations in [Aalen et al., 2008].

\subsection{Hazard, Inclusion, and the product-integral}

Let the model score $S$ be a random variable with the inclusion function $I(t) = P(S > s)$. We can assume that the inclusion function $I(s)$ is absolutely continuous. Let $f(s)$ be the density of $S$. The standard definition of the hazard rate $\alpha(s)$ of $S$ is the following with $ds$ being infinitesimally small. 

$$
I(s) = P(S > s) = 1 - F(s) = \int_s^{\infty}f(s)ds
$$

$$
\alpha(s) = \lim_{\Delta s \to 0}\frac{1}{\Delta s}P(s\leq S \leq s+\Delta s | S \geq s) = \frac{f(s)}{I(s)}
$$

The probability of a response occurring is in the immediate next score output. This way, alpha is obtainable from $S$.

$$
\alpha(s) = \frac{-I'(s)}{I(s)}
$$

Because $-f(s)$ is the derivative of I(s) we can rewrite the expression as the following. 

$$
\alpha(s) = \frac{-d}{ds}\log(I(s))
$$

$$
I(s) = \exp^{-\int_{0}^{s}\alpha(s)ds}
$$

\subsection{Markov Process}

Binary classification model output relates to a Markov process where the transition properties are score dependent. Let $X(s)$ be defined by the state space ${0,1}$ and by the transition intensity matrix.

$$
\bf{\alpha}(s) = 
\begin{bmatrix}
    -\alpha(s)       & \alpha(s)  \\
    0 & 0 
\end{bmatrix}
$$

State $1$ is thus absorbing, and the intensity of leaving state $0$ and entering state $1$ is $\alpha(s)$ at score $s$. 

\begin{align}
P(s) & = 
\begin{bmatrix}
    I(s)       & 1 - I(s)  \\
    0 & 1
\end{bmatrix}\nonumber\\
& = 
\begin{bmatrix}
    \exp^{(-\int_{0}^{s}\alpha(s)ds)}       & 1 - \exp^{(-\int_{0}^{s}\alpha(s)ds)}  \\
    0 & 1
\end{bmatrix}\nonumber\\
\end{align}

The Chapman-Kolmogorov equations for the forward equation is the following when the transition probabilities are absolutely continuous [Ross, 2014]. 

$$
\frac{\partial}{\partial s}{\bf P}(t,s) = {\bf P}(t,s) {\bf \alpha}(s)
$$

$$
\alpha(s) = \lim_{\Delta s \to 0}\frac{1}{\Delta s}({\bf P}(s, s+\Delta s) - {\bf I})
$$

The solution for the general case we apply the Chapman-Kolmogorov equations where $t = s_0 < s_1 < s_2 < \ldots < s_k = s$.

$$
{\bf P}(t,s) = {\bf P}(s_0,s_1){\bf P}(s_1,s_2)\cdots{\bf P}(s_{K-1},s_{K})
$$

When the lengths of the subintervals go to zero we arrive at the solution as a matrix product-integral. 

$$
{\bf P}(t,s) = \prod_{u\in(t,s]}\{{\bf I} + \alpha(u)du\}
$$

$$
I(s) = \prod_{u\leq s} (1-dA(u))
$$

When $A$ is absolutely continuous we write $dA(u) = \alpha(u)du$.

\begin{align}
I(s) &= \prod_{u\leq s} (1-dA(u)) = \prod_{u\leq s} (1-\alpha(u)du)\nonumber\\
&= \exp\{-\int_{u \leq s}\alpha(u)du\} = \exp^{-{\bf A}(s)}\nonumber
\end{align}

$$
A(s) = -\int_{0}^{s}\frac{dI(u)}{I(u^{-})}
$$

We simplify the version of $I(s)$ where we consider the conditional inclusion function $I(v|u) = P(S > v | S > u) = \frac{I(v)}{I(u)}$. This is the probability of a response occurring later than the score $v$ given that it has not occurred at score $u, v > u$. 

\subsection{Product Limit Estimator}

To find the product limit estimator curve, partition the ordered score data into intervals and use the multiplication rule for conditional probabilities to find the probability of inclusion. The probability of inclusion is the conditional probability that the response will occur with at least the score s given that the response has not received a lower score. We define $D(s)$ as the count of the number of responders up until score $s$ and $Y(s)$  as the count of the number of records at risk "just before" score $s$. The count of the number of records at risk are the number of responders remaining with a score equal to or greater than $s$. The standard estimator for the inclusion function is defined as follows for all values of $s$ in the range where there is data.

$$
I(s) = \prod_{k=1}^{K}I(s_k | s_{k-1})
$$

The product limit estimator, evaluated at a given score $s$, is approximately normally distributed in large samples. A standard $100 \times (1 - \alpha)\%$ confidence interval for $I(s)$ takes the form:

$$
\hat{I}(s) = \pm z_{1-\frac{\alpha}{2}}\hat{\tau}(s) 
$$

$$
\hat{\tau}^{2}(s) = \hat{I}(s)^{2} \sum_{I_j < s}\frac{1}{Y(S_j)^2}
$$

To derive the asymptotic distribution of the product limit estimator we establish the right hand side as a stochastic integral, therefore it is a mean zero martingale. 

$$
\frac{\hat{I}(s)}{I^{*}(s)} - 1 = - \int_{0}^{s} \frac{\hat{I}(u^{-})}{I^{*}(u)}d(\hat{A} - A^{*}(u))
$$

$$
I^{*}(s) = \prod_{u\leq s}\{1 - dA^{*}(u)\}
$$

This relation establishes the product limit estimator as a martingale. Using this principle, it is proven that the product limit estimator is almost unbiased, uniformly consistent, and normally distributed.

For all values of s beyond the largest observation score or before the smallest observation score the estimator is not well defined. The product limit estimator is a step function with jumps at the observed responder scores. The observed responders at score s and the censored observations just prior to score s determine the size of the jumps in the step function. 

$$
\hat{I}(s) = 
\begin{cases}
    1,& \text{if } s < s_1\\
    \prod_{s_i \leq s}[1-\frac{d_i}{Y_i}],              & \text{if } s_i \leq s
\end{cases}
$$

The variance of the product limit estimator is estimated by the following. 

$$
\hat{V}[\hat{I}(s)] = \hat{I}(s)^{2}\sum_{s_i \leq s}\frac{d_i}{Y_i (Y_i -d_i)} 
$$
The cumulative hazard has a unique relationship with the product limit estimator, and defined up the largest observed time as follows:

$$
\Tilde{A}(s) = 
\begin{cases}
    0,& \text{if } s \leq s_1\\
    \sum_{s_i \leq s}1-\frac{d_i}{Y_i},              & \text{if } s_i \leq s
\end{cases}
$$

$$
\sigma^{2}_{A}(s) = \sum_{s_i \leq s}\frac{d_i}{Y_i^{2}}
$$

\subsection{Cox Proportional Hazards Regression Model}

Explainability of the effects of the model covariates can be approximated through CPH regression model. The regression aims to predict the distribution of the score to response from a set of covariates. Covariates can be binary categorical or continuous and can also be time dependent. Time dependent features are incorporated as additional observations in the data with a censored response. The theoretical derivation remains the same. 

$$
\alpha(s | {\bf Z}) = \alpha_0(s)c(\beta^{s}Z)
$$

$$
\alpha(s | {\bf Z}) = \alpha_0(s)\exp(\beta^{s}Z) = \alpha_0(s)\exp(\sum_{k=1}^{p}\beta_kZ_k)
$$

where $\alpha_0(s)$ is an arbitrary baseline hazard rate. The parametric form is only assumed for the covariate effect. The baseline hazard rate is non-parametric. $\alpha(s | Z )$ must be positive. The model is called proportional hazards model because two observations with covariate values ${\bf Z}$ and ${\bf Z}^{*}$ have a ratio of hazard rates as:

$$
\frac{\alpha(s | {\bf Z})}{\alpha(s | {\bf Z}^{*})} = \frac{\alpha_0(s)\exp(\sum_{k=1}^{p}\beta_kZ_k)}{\alpha_0(s)\exp(\sum_{k=1}^{p}\beta_kZ_k^{*})} = \exp(\sum_{k=1}^p\beta_k(Z_k-Z_k^{*}))
$$

The hazard rates are proportional. If only one covariate, $Z_1$, differs and is binary categorical while all other covariates remain the same between ${\bf Z}$ and $\bf Z^{*}$, the proportional hazards become:

$$
\frac{\alpha(s | {\bf Z})}{\alpha(s | {\bf Z^{*}})} = \exp(\beta_1)
$$

The likelihood of the Beta vector is as follows:

$$
L_1(\beta) = \prod_{i=1}^{D} \frac{\exp(\beta_s s_i)}{\sum_{j\in R_i}\exp(\beta_s Z_j)^{d_i}}
$$

\subsection{Additive Cox Proportional Hazards Regression Models}

The CPH model incorporates time dependent features. An additional modeling step is able to fit score dependent coefficients using a generalized additive model of the CPH regression. Estimation of the additive nonparametric model focuses on the cumulative regression functions 

$$
B_q(s) = \int_0^s \beta_q(u)du
$$

where the estimation is performed at each response score by regressing for the observations at risk on their covariates. 

$$
\alpha_i(s) = Y_i(s)\{\beta_0(s) + \beta_1(s)x_{i1}(s) + \cdots + \beta_p(s)x_{ip}(s) \}
$$

Now $dD_i(s) = \alpha_i(s)ds + dM_i(s)$, so we may write

$$
dD_i(s) = Y_i(s)dB_0(s) + \sum_{j=1}^p Y_i(s)x_{ij}(s)dB_j(s) + dM_i(s)
$$

$i=1,2,\cdots,n$. This relation has the form of an ordinary linear regression model where the $dD_i(s)$ are the observations, the $Y_i(t)x_{ij}(s)$ are the covariates, the $dB_j(s)$ are the parameters to be estimated, and the $dM_i(s)$ are the random errors. 

\section{Approach}

We propose an approach that is inspired by survival analysis, a statistical field for measuring time to event data. The foundation of survival analysis uses a counting process that is derived from Markov processes, which generally define a random process with independent increments. The Markov process under study is the model score to observational response in classification models.

\subsection{Product Limit Estimator}
Each responder and non responder in the scored dataset is given a score output from the machine learning model. Scores from the classifier offer a ranking for which an observation $i$ is likely to be included as a response for category $k$. The score $S$ is an output from a machine learning model and can be interpreted as a probability or a utility for assigning an observation $i$ to category $k$. Each observation is either a responder or non responder and the score given to the observation is a random variable. 

The counting process uses order statistics from the score output file, where at each interval a confusion matrix summarizes the output of the model. The interval size can vary to be of equal length or calculated with each additional observation. The cumulative gains table measures the performance of the model for different score cutoffs. Only responders or censored observations are considered in the analysis while the non responders are truncated. The hazard, the product limit estimator incorporates the responders and censored observations to create the cumulative distribution function (CDF) of the probability inclusion [Kaplan and Meier, 1958]. The probability of inclusion is the conditional probability of a response at a score segment $j$ given the responder did not have a score greater than s, the score at segment $j$.

\subsection{Cox Proportional Hazards}
Explanations of the black box classification model can be found using the input variables as covariates in a Cox proportional hazards (CPH) regression model to explain the scores of the responder observations. The multiplicative hazards model quantifies the relationship between the black box model score to responder and a set of explanatory variables. The potential explanatory variables are the input variables used to train the classification model. The explanatory model can be used to find the baseline hazard rate, the hazard rate of an observation when all covariates are equal to zero. The effect of the covariates act multiplicatively on the baseline hazard and are assumed to be constant across all model scores.

\subsection{Additive Cox Proportional Hazards}
We take a step further to explain the black box classification model by using time dependent covariates in the CPH model. Covariates can be time dependent, as is the case for RNNs and time series data. In this model, the baseline hazard rate as well as the coefficients in the CPH model are dependent on the score given across observations over time. The coefficient is the excess risk at score $j$ for the corresponding covariate.  The effects of the covariates change over score and the time dependent coefficients are arbitrary regression functions. 

\subsection{Algorithm}

Algorithm 1 describes the method to derive score dependent explanations for classification models using time dependent data.

\begin{algorithm}[tb]
\caption{Score Dependent Model Explanation for Time Dependent Data}
\label{alg:algorithm}
\textbf{Input}: score points $s, s_1, ..., s_k$ such that $s_1 < ... < s_{k-1} < s_k $ and integers $x, x_1, ..., x_k$

\textbf{Output}: $\max L_(\beta)$
\begin{algorithmic}[1] %[1] enables line numbers
\IF{$x = 1$}
\STATE $I(s) = P(S > s)$
\STATE $\alpha(s) = \frac{-I'(s)}{I(s)}$
\STATE solve $\alpha(s | {\bf Z}) = \alpha_0(s)\exp(\sum_{k=1}^{p}\beta_kZ_k)$
\STATE solve $\alpha_i(s) = Y_i(s)\{\beta_0(s) + \beta_1(s)x_{i1}(s) + \cdots + \beta_p(s)x_{ip}(s) \}$
\STATE \textbf{return} $\hat{B_i}(s)$
\ENDIF
\end{algorithmic}
\end{algorithm}

\section{Experimental Evaluation}

\subsection{Data}

The data chosen is time to failure data of Blackblaze hard drives, which has published time series hard drive reliability statistics and insights based on hard drives in their data center. The data published are SMART (Self-Monitoring, Analysis and Reporting Technology) statistics used by HDD manufacturers to determine when disk failure is likely to happen. Manufacturers and models do not have a standard form of collecting data, so the paper uses a year of data for all of the Seagate Model ST4000DM000 hard drives. The Raw SMART 9 statistic is the count of hours the device has been powered on, which we use as the time variable in the data and normalize to years. All other SMART statistics are normalized to be between 0 and 1 from the raw data collected. Data is collected at a daily rate and a failure is recorded the day before the device fails or the last working day of the device ensuring the model is causal. 

\subsection{Architecture}

We structure a deep LSTM network to learn when a hard drive will fail. We structure the LSTM network to have three LSTM layers with 256 artificial neurons followed by two fully connected layers with 256 artificial neurons and 1 fully connected layer with 1 artificial neuron. The network uses a lookback window of 5 days of SMART statistics. There are 20 SMART statistics used as covariates as well as indicator statistics for each SMART feature if the feature is greater than zero. The model uses Relu activation functions and a drop out level of 0.2 for the LSTM layers. It uses a sigmoid activation function and l2 regularization of 0.002 for the fully connected layers. Adam optimizer is used with learning rate of 0.001.

\subsection{Training}
 The model was trained in 200 epochs with a batch size of 30 observations. The training classes fed into the RNN are balanced. The RNN received 0.7571 accuracy on the test dataset. The precision for the test dataset using the model is 0.9429 and recall is 0.5928, which is the highest accuracy achieved to our knowledge. 
 
\subsection{Explanation}
The Product Limit Estimator in Figure 1 is calculated by looking at all data from failed devices and their scores. 

\begin{figure}[]
\centering
\includegraphics[width=0.5\textwidth]{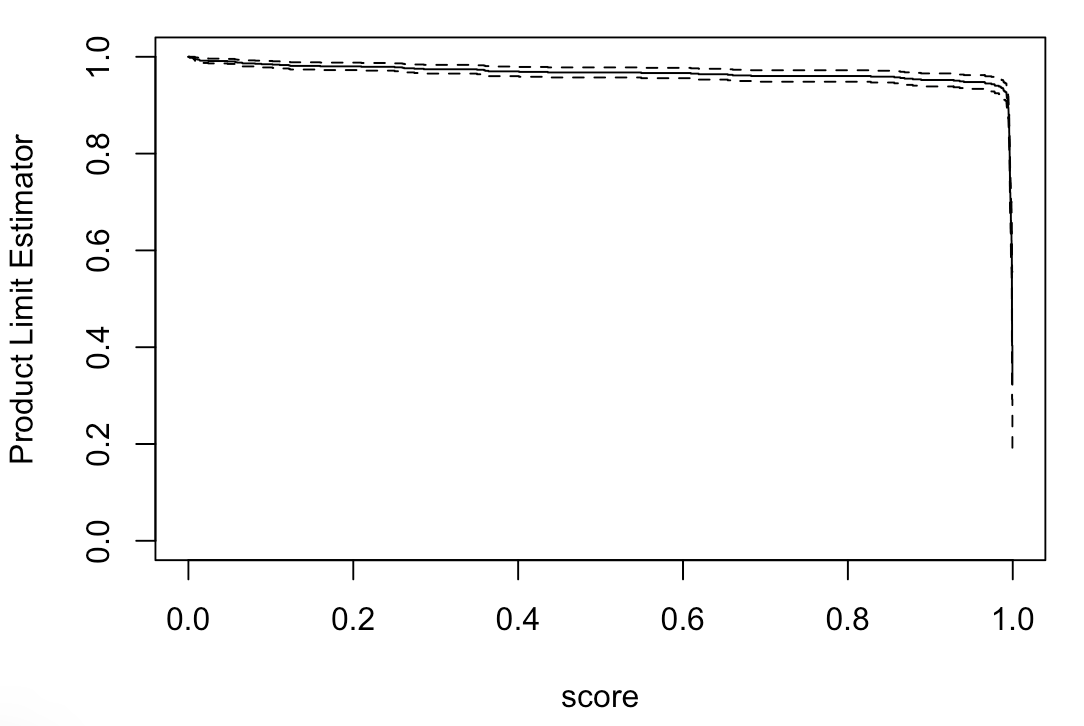}
\caption{Product Limit Estimator}
\label{fig:PLE}
\end{figure}

Our method of using time dependent data considers the data in the look back window as censored observations because it is unknown whether the device at these statistics would fail at the time the data was being collected. Covariates which are collinear are removed from the CPH analysis and only one of those covariates are used in the regression. We use forwards and backwards selection in the regression to determine the final significant covariates for the global explanation. Both of the significant covariates show an increase in probability of failure when the covariates are positive, shown in Figure 2. The results of the time dependent CPH model is in Table 1. 

\begin{figure}[]
\centering
\includegraphics[width=0.5\textwidth]{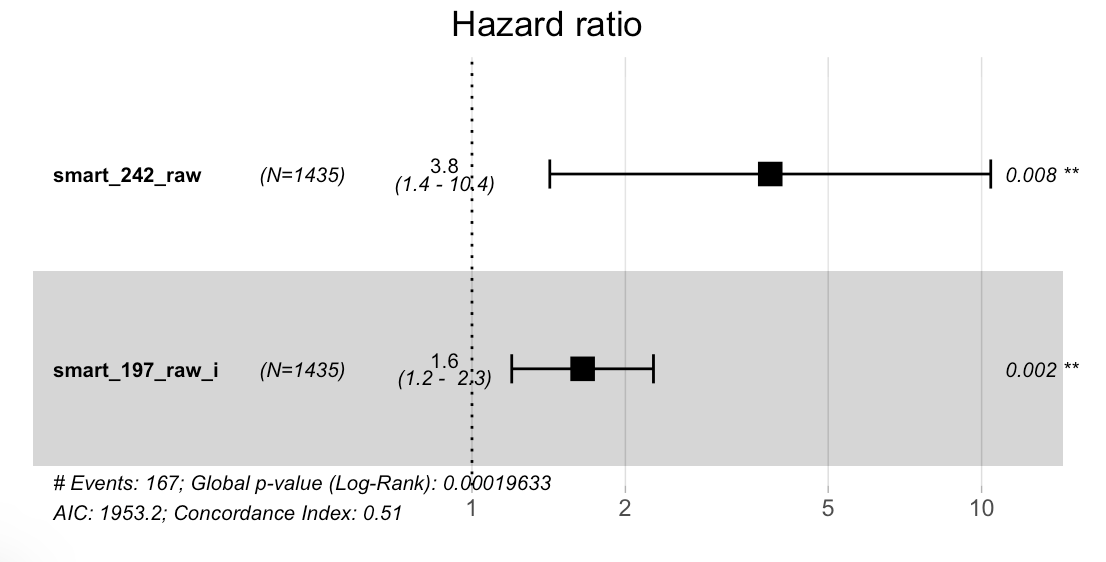}
\caption{CPH Model Coeffients}
\label{fig:CPH}
\end{figure}

The CPH explanation shows that there is a greater likelihood of failure for SMART 242, the total count of Logical Block Addressing (LBAs) reads, and the indicator for SMART 197, the count of unstable sectors waiting to be remapped because of unrecoverable read errors. The regression shows that a hard drive is 1.6484 times more likely to fail than the baseline hazard if it has a value greater than zero for SMART statistic 197 and 3.8486 times the normalized SMART 242 statistic times more likely than the baseline hazard rate.

\begin{table*}[h!]
\centering
\begin{tabular}{lrrrrrr}  
\toprule
Covariate  & Coeff & exp(Coeff) & se(Coeff) & z & P value & MSE Ratio\\
\midrule
SMART 197 i & 0.4998 & 1.6484 & 0.1634 & 3.059 & 0.00222 & $\texttt{+} 8.1\%$\\
SMART 242 & 1.3477 & 3.8486 & 0.5083 & 2.652 & 0.00801 & $\texttt{+} 3.9\%$\\
\bottomrule
\end{tabular}
\caption{CPH Explanation with Time Dependent Data}
\label{tab:booktabs}
\end{table*}

Although both significant covariates had a multiplicative factor for failure greater than 1 in the CPH regression model, the score dependent explanations offer more detail showing that SMART 242 in Figure 3 continues to lead to a greater probability of failure in any score region while SMART 197 i in Figure 4 leads to a greater probability of failure only for higher score regions while in lower score regions indicates a lower probability of failure than the baseline hazard rate. Performing sensitivity analysis on the features found the most variation in model performance by altering the values in SMART 184 End-to-End Error $\texttt{+} 23.1\%$ MSE Ratio and SMART 7 Seek Error Rate $\texttt{+} 22.1\%$ MSE Ratio.

\begin{figure}[]
\centering
\includegraphics[width=0.5\textwidth]{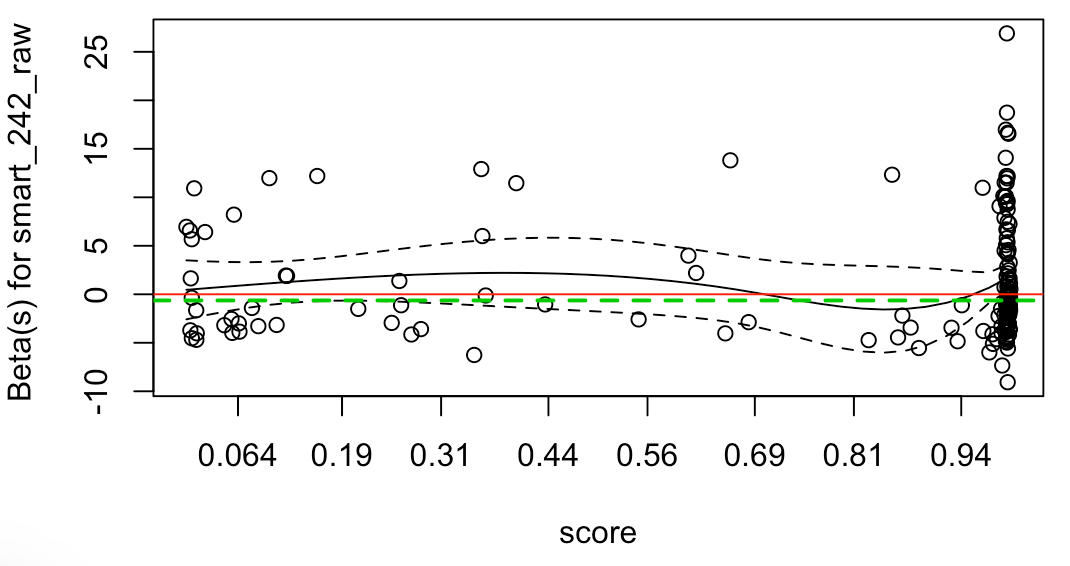}
\caption{Beta Coefficient SMART 242}
\label{fig:Beta242}
\end{figure}

\begin{figure}[]
\centering
\includegraphics[width=0.5\textwidth]{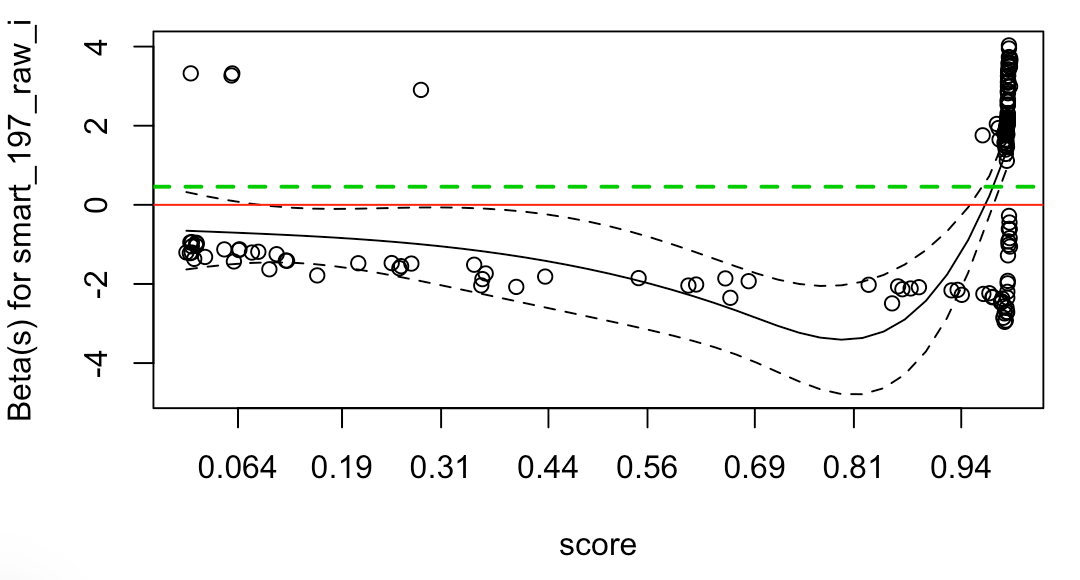}
\caption{Beta Coefficient SMART 197 i}
\label{fig:Beta197i}
\end{figure}

This explanation of features highlights the difficulty in predicting hard drive failures and demonstrates why we are able to receive high precision but low recall in the test dataset.

\section{Conclusion}

The method describes a non-parametric counting process to define the cumulative probability of a responder record occurring by a score segment. We show that a Markov process state space model can be applied to evaluate the stochastic process of observations over the time series classification model score. We formulate a new definition for the Recall curve as the cumulative probability of a responder being classified as a responder, a true positive. The explanations provided attributes the likelihood of response to feature inputs used in the black box model, even when the features are time series and in order dependent models such as recurrent neural networks. Finally, we present a novel method to use information from the time dependence of the features in the explanation and derive local score dependent explanations.

\vspace{-0.2cm}
\section*{Acknowledgments}
We would like to thank Blackblaze for providing the data for which without their contribution the analysis would not be possible. 
\vspace{-0.1cm}

\bibliographystyle{named}

\end{document}